# Biometric identification by means of hand geometry and a neural net classifier

Marcos Faundez-Zanuy, Guillermo Mar Navarro Mérida

Escola Universitària Politècnica de Mataró
Universitat Politècnica de Catalunya, BARCELONA (SPAIN)
`faundez@eupmt.es`
`http://www.eupmt.es/veu`

**Abstract**. This Paper describes a hand geometry biometric identification system. We have acquired a database of 22 people using a conventional document scanner. The experimental section consists of a study about the discrimination capability of different extracted features, and the identification rate using different classifiers based on neural networks.

## 1 Introduction

In recent years, hand geometry has become a very popular access control biometrics which has captured almost a quarter of the physical access control market [1]. Even if the fingerprint is most popular access system [2-4], the study of other biometric systems is interesting, because the vulnerability of a biometric system [5] can be improved using some kind of data fusion [6] between different biometric traits. This is a key point in order to popularize biometric systems [7], in addition to privacy issues [8].

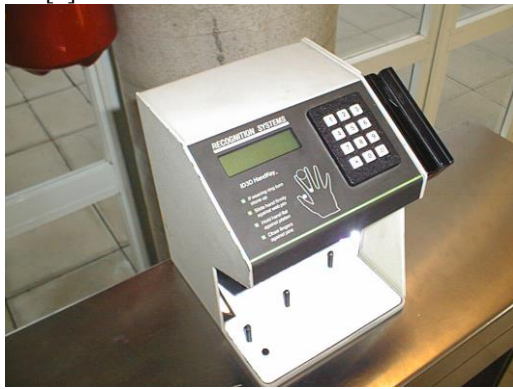

**Fig. 1.** Commercial three-dimensional scanner.

Although some commercial systems, such us the system shown in figure 1 rely on a three-dimensional profile of the hand, in this paper we study a system based on two dimensional profiles. Although three dimensional devices provide more information than two dimensional ones, they require a more expensive and voluminous hardware.

A two-dimensional profile of a hand can be get using a simple document scanner, which can be purchased for less than 100 USD. Another possibility is the use of a digital camera, whose cost is being dramatically reduced in the last years.

In our system, we have decided to use a conventional scanner instead of a digital photo camera, because it is easier to operate, and cheaper. On the other hand, although a digital camera is extremely fast in taking a photo, the last generation scanners (such as EPSON 4870 Photo perfection) are able to capture a DIN A4 size colour document (24 bit) at a 150 dpi resolution in less than 15 seconds when using the USB 2 port, which is a quite reasonable time.

This paper can be summarized in three main parts: section two describes a database which has been specially acquired for this work. In section three, we describe the pre-processing and we study the discrimination capability of several measurements on the sensed data. Section four provides experimental results on identification rates using neural net classifiers.

## 2 Database

We have acquired a database of 22 people, and 10 different acquisitions per person. If some acquisition has not let to extract some of the parameters described in the next section, this capture has been rejected and replaced by a new one. Figure 2 shows an example of defective acquisitions and the reason.

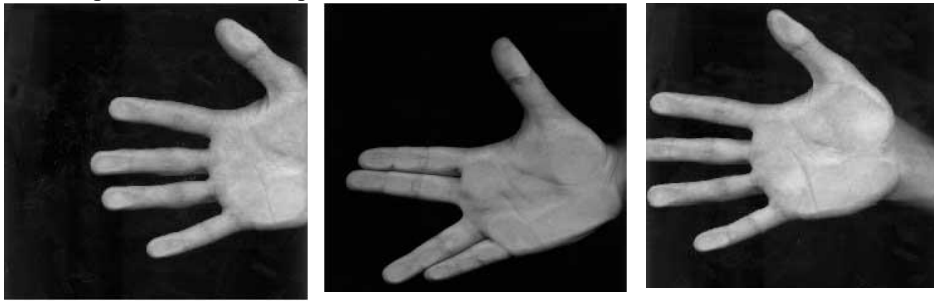

**Fig. 2.** Example of defective acquisitions. The first one is defective because it is cut on the base. In the second one, some fingers are joined. In the third one, one finger is cut.

The database has been stored in bmp format using 8 bits per pixel (256 gray levels), a resolution of 100 dpi, and an image size of 216x240 mm. Higher resolutions would imply more details but also more computational time in order to process a hand image. In our preliminary experiments we have found that 100 dpi offers a good compromise. Obviously this resolution is insufficient for other related applications such as palm print, which is analogous to fingerprint recognition, but using the ridge and valley pattern of the hand skin. Thus, the system will rely on the silhouette of the hand and will ignore other details such as fingerprints, lines, scars and color. For this rea-son, the first step of the pre-processing described in the next section will be a binarization and a contour extraction. Although this procedure discards useful information for discrimination, it also alleviates other problems, such as the perspiration of the skin which blots the thin details of the image. Figure 3 shows an example of this phenomenon.

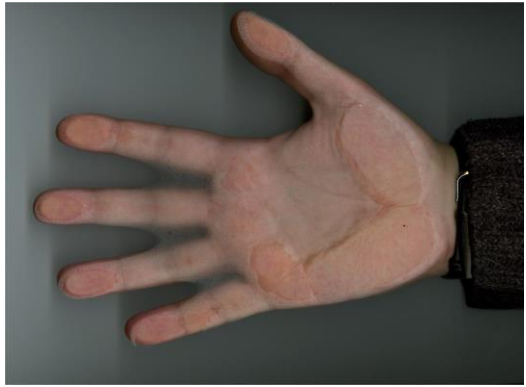

**Fig. 3.** Example of hand acquisition at 150 dpi and 24 bit per pixel (color image), with perspiration problem. This problem can be neglected after the binarization step.

## 3 Feature extraction

### 3.1 Pre-processing algorithm

Figure 4 shows a block diagram of the pre-processing algorithm.

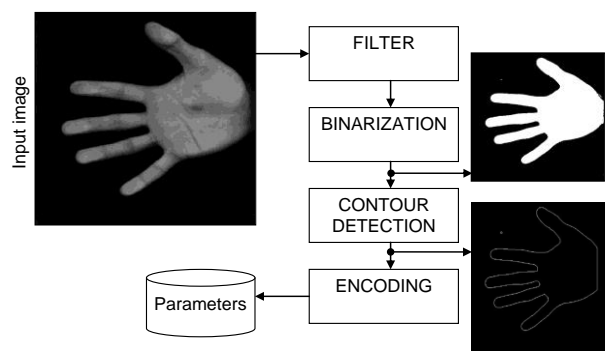

**Fig. 4.** Block diagram for the proposed pre-processing scheme.

The description of each block is the following:

**Filter**

We apply a low-pass filtering in order to remove spurious noise.

**Binarization**

The goal is the conversion from an image $I(x,y)$ at 8 bit per pixel to a monochrome image $I'(x,y)$ (1 bit per pixel. "0"=black, "1"=white), applying a threshold:

$$I'(x,y) = \begin{cases} 1 & if\ I(x,y) \geq threshold \\ 0 & otherwise \end{cases} \quad (1)$$

We use threshold=0.07

**Contour detection**

The goal is to find the limits between the hand and the background. For this purpose the algorithm detects the intensity changes, and marks a closed set of one pixel wide and length the perimeter of the hand. Edge points can be thought of as pixel locations of abrupt grey-level change. For example it can be defined an edge point in binary images as black pixels with at least one white nearest neighbour. We use the Laplacian of Gaussian method, which finds edges by looking for zero crossings after filtering the image with a Laplacian of Gaussian filter.

**Coding**

This step reduces the amount of information. We translate a bmp file to a text file that contains the contour description. The encoding algorithm consists of a chain code. In chain coding the direction vectors between successive boundary pixels are encoded. Figure 5 shows our code, which employs 8 possible directions and can be coded by 3-bit code words.

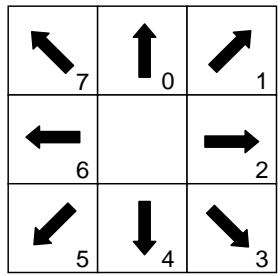 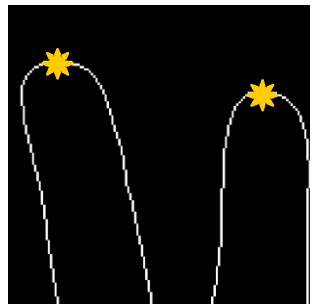

**Fig. 5.** Contour coding algorithm.   **Fig. 6.** Maximum of first and middle fingers.

Once upon the chain code is obtained, the perimeter can be easily computed: for each segment, an even code implies +1 and an odd code $+\sqrt{2}$ units. the beginnings and ends of the fingers and wrist are found looking for minimum and maximum values in the chain code.

The finger limits (base and maximum height) are detected in the middle of a region with a "5" and "3"·code. Figure 6 shows, for example, the maximum of the first and middle fingers.

### 3.2    Proposed features

Using the result of the previous section as input, we propose the following measurements (see figure 7):

1. Thumb finger length.
2. First finger length.
3. Middle finger length.
4. Ring finger length.
5. Little finger length.
6. Wrist length.
7. Thumb base width.
8. First finger width.
9. Middle finger width.
10. Ring finger width.
11. Little finger width.
12. Hand perimeter.
13. Hand surface.

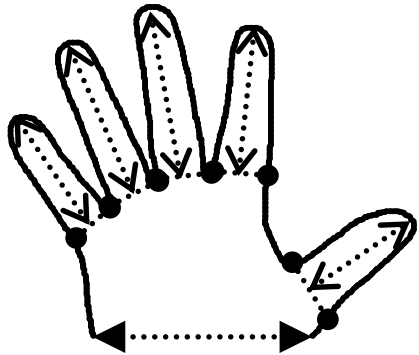

Fig. 7. Measured features.

Some of these features have been removed in the experiments due to their low discrimination capability. Our experiments have revealed that results are improved deleting features 1, 6, 7 and 13. Thus, we will select the remaining nine features per image.

## 4    Experimental results and Conclusions

### 4.1    Conditions of the experiments

Our results have been obtained with the database described in section 2, the preprocessing of section 3, and the selected parameters of section 4, in the following situation: 22 persons, images 1 to 5 for training, and images 6 to 10 for testing.

### 4.2    Nearest neighbour classifier

We obtain one model from each training image. During testing each input image is compared against all the models inside the database (22x5=110 in our case) and the model close to the input image (using Mean Square Error criterion) indicates the recognized person.

In our experiments, we are making for each user, all other users' samples as impostor test samples, so we finally have, $N=22\times5$ (client)+$22\times21\times5$ (impostors)=2420 different tests. We have used two different distance measures:

$$MSE(\vec{x},\vec{y})=\sum_{i=1}^{P}(x_i-y_i)^2 \qquad (2)$$

$$MAD(\vec{x},\vec{y})=\sum_{i=1}^{P}|x_i-y_i| \qquad (3)$$

Where $P$ is the vector dimension.

### 4.3 Multi-Layer Perceptron classifier trained in a discriminative mode

We have trained a Multi-Layer Perceptron (MLP) [9] as discriminative classifier in the following fashion: when the input data belongs to a genuine person, the output (target of the NNET) is fixed to 1. When the input is an impostor person, the output is fixed to –1. Figure 8 shows the neural network architecture. We have used a MLP with 30 neurons in the hidden layer, trained with the Levenberg-Marquardt algorithm, which computes the approximate Hessian matrix, because it is faster and achieves better results than the classical back-propagation algorithm. We have trained the neural network for 10 epochs (50 epochs when using regularization). We also apply a multi-start algorithm and select the best result.

The input signal has been fitted to a [–1, 1] range in each component.

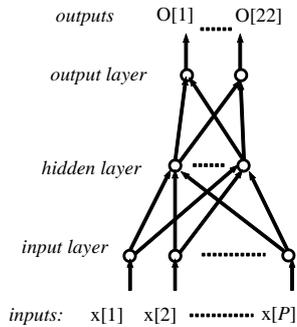

**Fig. 8.** Multi-Layer Perceptron architecture.

One of the problems that occur during neural network training is called *overfitting*. The error on the training set is driven to a very small value, but when new data is presented to the network the error is large. The network has memorized the training examples, but it has not learned to generalize to new situations. The adopted solution to the overfitting problem has been the use of regularization. The regularization involves modifying the performance function, which is normally chosen to be the sum of squares of the network errors on the training set. So, this technique helps take the mystery out of how to pick the number of neurons in a network and consistently leads to good networks that are not overtrained. The classical Mean Square Error (MSE) implies the computation of (4):

$$MSE = \frac{1}{N}\sum_{i=1}^{P}(t_i - a_i)^2 \qquad (4)$$

Where $t$, $a$ are the $P$ dimensional vectors of the test input and the model, respectively. The regularization uses the following measure (5):

$$MSEREG = \gamma MSE + (1-\gamma)\frac{1}{n}\sum_{j=1}^{n}w_j^2 \qquad (5)$$

Thus, it includes one term proportional to the modulus of the weights of the neural net.

In addition, there is another important topic: the random initialization. We have studied two strategies:
a) To pick up the best random initialization (the initialization which gives the higher identification rate)
b) A committee of neural networks, which combines the outputs of several MLP, each one trained with a different initialization.

### 4.4 Radial Basis Function classifier trained in a discriminative mode

We have trained a Radial Basis Function (RBF) in a similar fashion than MLP of previous section. Figure 9 shows the architecture. Taking into account that a RBF is faster to train, we have worked out a exhaustive study varying the number of centres. Figure 10 shows the identification rate as function of the number of centres. It can be seen that the maximum value is 89.09%, which is achieved using 50 centres.

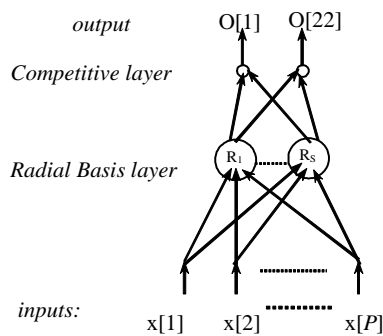

**Fig. 9.** Radial Basis Function architecture.

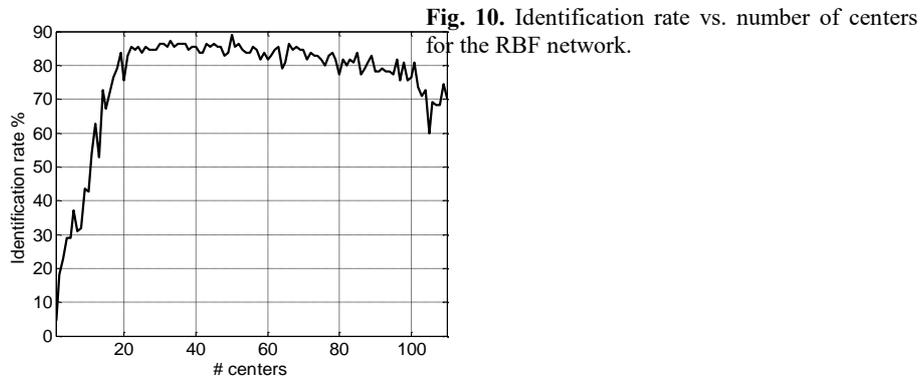

**Fig. 10.** Identification rate vs. number of centers for the RBF network.

### 4.5 Experimental Results

Table 1 compares the neural net results with the classical nearest neighbour classifier with two different distance measures. It can be appreciated that the neural networks outperform the Nearest Neighbour classifier.

Although hand-geometry does not offer the good results of fingerprint biometric recognition [10-11], it can be more accepted by the uses, because fingerprint are more related to police, and criminal records.

**Table 1.** Comparison between different Classifiers

| Classifier | Identification rate (%) |
|---|---|
| Nearest Neighbor (MAD) | 64,55% |
| Nearest Neighbor (MSE) | 73,64% |
| Multi-Layer Perceptron (MSE, 10 epoch) | 91,82% |
| Multi-Layer Perceptron (MSEREG, 50 epoch) | 92,73% |
| MLP committee 3 nets (MSE, 10 epoch) | 93,64% |
| MLP committee 3 nets (MSEREG, 50 epoch) | 93,64% |
| Radial Basis Function | 90% |

Marcos Faundez-Zanuy and Guillermo Mar Navarro Mérida. 2005. Biometric identification by means of hand geometry and a neural net classifier. In Proceedings of the 8th international conference on Artificial Neural Networks: computational Intelligence and Bioinspired Systems (IWANN'05). Springer-Verlag, Berlin, Heidelberg, 1172–1179. DOI:https://doi.org/10.1007/11494669_144

## Acknowledgement

This work has been supported by FEDER and the Spanish grant MCYT TIC2003-08382-C05-02.